# Dynamic Jointrees

**Adnan Darwiche**
Department of Mathematics
American University of Beirut
PO Box 11 - 236
Beirut, Lebanon
*darwiche@aub.edu.lb*

## Abstract

It is well known that one can ignore parts of a belief network when computing answers to certain probabilistic queries. It is also well known that the ignorable parts (if any) depend on the specific query of interest and, therefore, may change as the query changes. Algorithms based on jointrees, however, do not seem to take computational advantage of these facts given that they typically construct jointrees for worst-case queries; that is, queries for which every part of the belief network is considered relevant. To address this limitation, we propose in this paper a method for reconfiguring jointrees dynamically as the query changes. The reconfiguration process aims at maintaining a jointree which corresponds to the underlying belief network after it has been pruned given the current query. Our reconfiguration method is marked by three characteristics: (a) it is based on a non-classical definition of jointrees; (b) it is relatively efficient; and (c) it can reuse some of the computations performed before a jointree is reconfigured. We present preliminary experimental results which demonstrate significant savings over using static jointrees when query changes are considerable.

## 1 Introduction

There is a number of algorithms for exact inference in belief networks, but the ones based on jointrees seem to be the most dominant [3, 4, 5]. According to these algorithms, the structure of a belief network is converted into a jointree which is then used as the basis for computing posterior probabilities given evidence.

One of the main drawbacks of jointree algorithms, however, is that they prepare for worst-case queries.

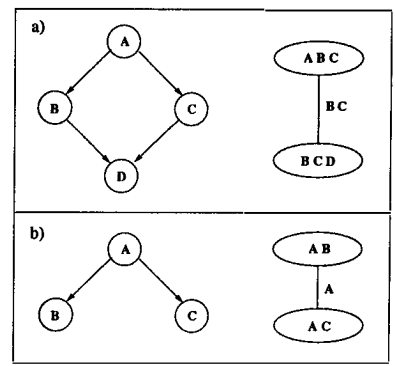

Figure 1: Belief networks and respective jointrees.

That is, these algorithms typically build jointrees under the assumption that every node in the belief network can be observed or updated (its posterior probability computed). This means that the whole structure of the belief network is involved in computing a jointree which may, in certain situations, turn out to be an overkill. This happens when only part of the belief network turns out to be relevant to a given query, therefore, permitting inference with respect to a much simpler jointree.

Consider the belief network and its corresponding jointree shown in Figure 1(a) for an example. This jointree is built for the worst-case: any node can be observed and any node can be updated. Suppose, however, that in reality only Node $B$ is observed and only Node $C$ needs to be updated. In such a case, the pruned belief network and its corresponding jointree shown in Figure 1(b) will suffice for handling the situation.

Adopting the jointree in Figure 1(b) will clearly involve less work, but the problem is that one may not know upfront that the query of interest is $\mathbf{Pr}(c \mid b)$. Therefore, one is forced to build the jointree of Figure 1(a) in the first place to prepare for the worst case.

One suggestion, however, is to postpone the construction of the jointree until a specific query material-izes. Once the query is known, the belief network can be pruned and a jointree can be constructed for the pruned belief network.

There are two main problems with such a proposal however. First, building a jointree is costly. Therefore, building a jointree each time the query changes may not be cost effective. Second, even if building a jointree can be done efficiently, changing the jointree each time the query changes means that we are tossing away the results of any computations performed thus far. But the reuse of such computations has been crucial for the performance of existing jointree algorithms. Therefore, building a new jointree each time the query changes could undermine any savings that are expected from having a better jointree.

It appears, therefore, that any successful proposal for reconfiguring jointrees must satisfy at least two properties. First, it must provide a method for reconfiguring jointrees efficiently. Second, it must provide a mechanism for reusing the results of computations performed before the jointree is reconfigured.

In this paper, we propose a method for dynamically reconfiguring jointrees in response to query changes. The method is shown to satisfy the above two properties and is inspired by a non-classical definition of jointrees, which is the subject of Section 2. Based on this non-classical definition, we propose a method for building jointrees in Section 3 and study its properties. In particular, we show how jointrees constructed using this method can be reconfigured in response to query changes. We then turn in Section 4 to a theorem which allows for the efficient reconfiguration of jointrees using our method. Section 5 is then dedicated to another important theorem which explicates conditions under which computations performed before reconfiguring a jointree remain valid after reconfiguration. Experimental results are then given in Section 6 where we demonstrate significant savings using this new method when query changes are considerable. We finally close in Section 7 with some concluding remarks.

Proofs of all theorems can be found in [1].

## 2  A Non-Classical Definition of Jointrees

We review in this section the standard definition of a jointree and then propose an alternative definition which will underly our formal developments in this paper. We start first with some preliminary definitions.

A *graph* is a pair $(\mathbf{V}, \mathbf{A})$ where $\mathbf{V}$ is a finite set of *nodes* and $\mathbf{A}$ is a subset of $\mathbf{V} \times \mathbf{V}$ known as *edges*. A *labeled graph* is a triple $(\mathbf{V}, \mathbf{A}, \mathbf{L})$ where $(\mathbf{V}, \mathbf{A})$ is a graph and $\mathbf{L}$ is a function that maps each node $i$ in $\mathbf{V}$ to a label $\mathbf{L}_i$. *Directed acyclic graphs (dags)*, *undirected graphs*, and *trees* are special cases of graphs obtained by imposing the usual conditions on the set of edges $\mathbf{A}$. The *family* of a node in a dag $\mathcal{G}$ is the set containing the node and its parents in $\mathcal{G}$.

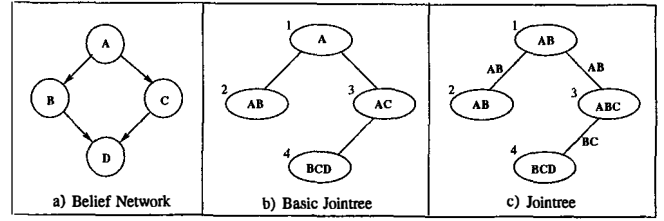

Figure 2: A belief network, a corresponding basic jointree, and a classical jointree induced by the basic jointree. In (b) and (c), the label of each node is shown inside it.

**Definition 1** *A jointree for a directed acyclic graph $\mathcal{G}$ is a labeled tree $\mathcal{T} = (\mathbf{V}, \mathbf{A}, \mathcal{C})$ where*

1. *Each label $\mathcal{C}_i$ is called a clique.*

2. *Each family in $\mathcal{G}$ is contained in some clique $\mathcal{C}_i$.*

3. *If a node belongs to two cliques $\mathcal{C}_i$ and $\mathcal{C}_j$, it must also belong to every clique $\mathcal{C}_k$ where $k$ is a node on the path connecting nodes $i$ and $j$ in $\mathcal{T}$.*

The *separator* of edge $(i, j)$ in $\mathbf{A}$ is defined as $\mathcal{S}_{ij} \stackrel{\text{def}}{=} \mathcal{C}_i \cap \mathcal{C}_j$.

The jointree in Figure 1(a) has two nodes with the associated cliques being $\{A, B, C\}$ and $\{B, C, D\}$. It also has one separator $\{B, C\}$.

We now present an alternative definition of jointrees, which has inspired our proposal for reconfiguring jointrees in response to a query change. Simply stated, a jointree for dag $\mathcal{G}$ is nothing but an aggregation of the families of $\mathcal{G}$ into groups which are connected in the form of a tree.

**Definition 2** *A basic jointree for a directed acyclic graph $\mathcal{G}$ is a labeled tree $\mathcal{T} = (\mathbf{V}, \mathbf{A}, \mathcal{H})$ where*

1. *Each label $\mathcal{H}_i$ is called a hypernode.*

2. *Each family in $\mathcal{G}$ is contained in some hypernode $\mathcal{H}_i$.*

3. *Each hypernode $\mathcal{H}_i$ is the union of some families in $\mathcal{G}$.*

Figure 2(a) depicts a dag and Figure 2(b) depicts a corresponding basic jointree with four hypernodes: $\mathcal{H}_1 = \{A\}$, $\mathcal{H}_2 = \{A, B\}$, $\mathcal{H}_3 = \{A, C\}$ and $\mathcal{H}_4 = \{B, C, D\}$. Here, each hypernode contains a single family of the dag, which creates a one-to-one correspondence between the families of the dag and the hypernodes of its basic jointree.

Given a labeled tree $\mathcal{T} = (\mathbf{V}, \mathbf{A}, \mathbf{L})$ and an edge $(i, j)$ in $\mathbf{A}$, we will use $\mathbf{L}_{ij}$ to denote the union of all labels $\mathbf{L}_k$ where $k$ is a node on the $i$-side of the edge $(i,j)$. For example, in Figure 2(b), $\mathcal{H}_{34} = \mathcal{H}_1 \cup \mathcal{H}_2 \cup \mathcal{H}_3 = \{A, B, C\}$ and $\mathcal{H}_{43} = \mathcal{H}_4 = \{B, C, D\}$.

Note that Definition 2 of a basic jointree does not mention separators or cliques. These are derivative notions:

**Definition 3** *Let $(\mathbf{V}, \mathbf{A}, \mathcal{H})$ be a basic jointree. The <u>separator</u> associated with edge $(i,j)$ in $\mathbf{A}$ is defined as follows:*

$$\mathcal{S}_{ij} \stackrel{def}{=} \mathcal{H}_{ij} \cap \mathcal{H}_{ji}.$$

*Moreover, the <u>clique</u> associated with node $i$ in $\mathbf{V}$ is defined as follows:*

$$\mathcal{C}_i \stackrel{def}{=} \mathcal{H}_i \cup \bigcup_j \mathcal{S}_{ij}.$$

That is, the separator associated with an edge in a basic jointree contains nodes which are shared by families on opposite sides of the edge. In Figure 2(b), $\mathcal{H}_{34} = \{A, B, C\}$ and $\mathcal{H}_{43} = \{B, C, D\}$. Therefore, $\mathcal{S}_{34} = \{B, C\}$ which are nothing but the nodes shared by families on opposite sides of the edge $(3, 4)$.

The clique associated with a node in a basic jointree contains its hypernode and all adjacent separators. In Figure 2, $\mathcal{H}_3 = \{A, C\}$, $\mathcal{S}_{13} = \{A, B\}$ and $\mathcal{S}_{43} = \{B, C\}$. Therefore, $\mathcal{C}_3 = \mathcal{H}_3 \cup \mathcal{S}_{13} \cup \mathcal{S}_{43} = \{A, B, C\}$.

**Theorem 1** *Let $\mathcal{T} = (\mathbf{V}, \mathbf{A}, \mathcal{H})$ be a basic jointree for dag $\mathcal{G}$ and let $\mathcal{C}_i$ be the clique associated with node $i$ in $\mathbf{V}$. The labeled tree $(\mathbf{V}, \mathbf{A}, \mathcal{C})$ is then a jointree for dag $\mathcal{G}$. Moreover, for every edge $(i,j)$ in $\mathbf{A}$, we have $\mathcal{C}_i \cap \mathcal{C}_j = \mathcal{H}_{ij} \cap \mathcal{H}_{ji}$.*

This theorem asserts two important results. First, that we can covert any basic jointree into a classical jointree by simply converting each of its hypernodes into a clique as given by Definition 3. Second, that the separators of a basic jointree and those of its induced jointree are equal. Figure 2(c) depicts the jointree induced by the basic jointree in Figure 2(b).[1]

---

[1] From the classical viewpoint, cliques $\mathcal{C}_1$ and $\mathcal{C}_2$ in Figure 2(c) are redundant and should be eliminated. However, we shall keep them because we would like to establish a one-to-one correspondence between the nodes of a belief network and the cliques of its jointree. This will be discussed further in the following section.

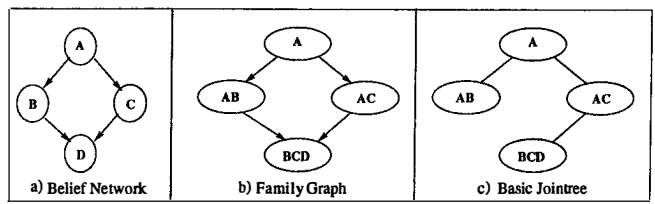

Figure 3: A belief network, its family graph and a spanning tree of the graph. In (b) and (c), the label of each node is shown inside it.

Note that in the standard definition of jointrees, cliques are primary objects and separator are secondary ones. Specifically, one defines a jointree by imposing a property on cliques and then defines separator as a further detail needed by jointree algorithms. In our alternative definition of jointrees, however, both cliques and separators are derivative objects. *The essence of a jointree according to our definition is an aggregation of families into a tree structure, leading to what we call a basic jointree.* Once such an aggregation is committed to, separators and cliques follow as derivative objects that facilitate computations. Even then, however, separators are primary and cliques are secondary.

Theorem 1, therefore, suggests a non-classical definition of jointrees which promotes two key points. First, that the defining characteristic of a jointree is an *aggregation* of families into groups and a *connection* of these groups into a tree. Second, that the separator associated with an edge is nothing but the intersection of families on opposite sides of the edge. We shall see in the following section how this viewpoint of jointrees can be exploited to reconfigure jointrees in response to a query change.

## 3  A Non-Classical Method for Building Jointrees

According to Theorem 1, one can construct a jointree $\mathcal{T}$ for dag $\mathcal{G}$ by simply constructing a basic jointree for $\mathcal{G}$ according to Definition 2 and then converting each of its hypernodes into a clique as given by Definition 3. The second step is deterministic, but the first step can be realized using a number of methods. We shall adopt a specific method in the rest of this paper for the sake of concreteness. Specifically, we will construct a basic jointree for a dag by simply connecting its families into a tree structure.

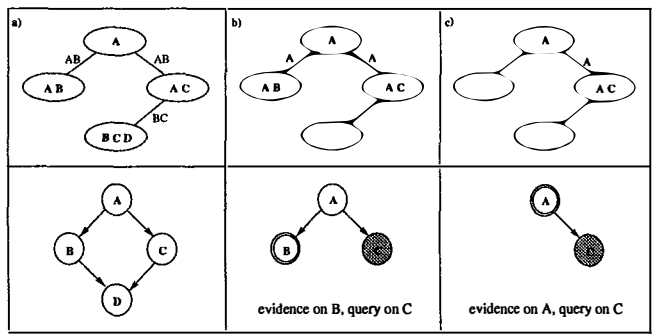

Figure 4: Three belief networks with corresponding basic jointrees. Double circles represent evidence nodes, filled circles represent nodes to be updated.

**Definition 4** *Let $\mathcal{G} = (\mathbf{V}, \mathbf{A})$ be a dag. The family graph of $\mathcal{G}$ is a labeled dag $(\mathbf{V}, \mathbf{A}, \mathcal{F})$ where $\mathcal{F}_i$ is the family of node $i$ in $\mathcal{G}$.*

Each spanning tree of this family graph is then a basic jointree.

**Theorem 2** *Let $\mathcal{G}'$ be the family graph of dag $\mathcal{G}$ and let $\mathcal{T}$ be a spanning tree of $\mathcal{G}'$. Then $\mathcal{T}$ is a basic jointree for $\mathcal{G}$.*

Figure 3 depicts a belief network, its family graph and a basic jointree which corresponds to a spanning tree of the family graph.[2] According to our method then, constructing a basic jointree is just a matter of deleting enough arcs from the family graph until the graph becomes singly connected.

Realize that once a basic jointree is constructed, separators and cliques are determined uniquely given Definition 3. We shall therefore speak mostly about the basic jointree, leaving separators and cliques implicit.

*Now here's the basic observation underlying our proposal for reconfiguring jointrees in response to a query change.* Consider the belief network and its basic jointree shown in Figure 4(a). This basic jointree can be viewed as preparing for the worst-case: it can be used for computing the posterior probability of any node given evidence about any other node. Suppose now that we have evidence about Node $B$ and we want to update Node $C$. We can, of course, use the basic jointree in Figure 4(a) to handle this query, but this amounts to performing inference with respect to the full belief network. As we have observed earlier, however, the simpler, pruned belief network in Figure 4(b) is sufficient for handling this query. Therefore,

we will reconfigure the basic jointree in Figure 4(a) so it becomes a basic jointree for the pruned network in Figure 4(b).[3] This can be achieved by simply removing the family of Node $D$, given that it was pruned, from its corresponding hypernode, leading to the basic jointree in Figure 4(b). This simpler, basic jointree is indeed a basic jointree for the pruned network in Figure 4(b) (according to Definition 2), and, therefore, can be used to answer the above query.

The simple method we used in the previous example is valid in general. That is, all we have to do is remove the family of each pruned node from its corresponding hypernode. We shall formalize this method now, but only after formalizing the notions of a query and that of pruning a belief network given a query.

**Definition 5** *Let $\mathbf{E}$ and $\mathbf{Q}$ be two sets of nodes in dag $\mathcal{G}$ and let $\mathbf{e}$ be an instantiation of nodes $\mathbf{E}$.[4] The pair $(\mathbf{e}, \mathbf{Q})$ is called a query for dag $\mathcal{G}$.*

The intuition here is that $\mathbf{e}$ represents evidence about nodes $\mathbf{E}$ and $\mathbf{Q}$ represents nodes whose posterior probabilities must be computed.

Given a belief network and a query $(\mathbf{e}, \mathbf{Q})$, not all nodes of the network may be relevant to the computation. In particular, any leaf node in the network which does not belong to either $\mathbf{E}$ or $\mathbf{Q}$ can be removed from the network. When this pruning operation is applied recursively, a large portion of the belief network may be pruned, which in turn may lead to a much simpler jointree. Our goal, of course, is to reconfigure the originally constructed jointree so it corresponds to this pruned belief network. We put two constraints on ourselves, however: efficiency of reconfiguration, and reuse of previously performed computations. We will address these two constraints in the following two sections, but first we formalize the reconfiguration process.

**Definition 6** *Let $\mathcal{G}$ be a dag and let $q = (\mathbf{e}, \mathbf{Q})$ be a query for $\mathcal{G}$. The pruning of $\mathcal{G}$ given $q$, denoted $\mathcal{G}^q$, is the dag which results from successively removing leaf nodes from $\mathcal{G}$ if they are not in $\mathbf{E} \cup \mathbf{Q}$.*

This is a standard definition of pruning where removed leaf nodes are known as barren-nodes.

---

[2]It appears that our method for constructing jointrees can be justified using the transformation approach given in [2]. In particular, computing a spanning tree of the family graph can be viewed as successively applying the *collapse transformation* to the trivial *cluster graph* as defined in [2].

[3]We have two choices when trying to perform inference with respect to the pruned belief network in Figure 4(b): We can directly compute a basic jointree for the network or we can reconfigure the basic jointree in Figure 4(a) for that purpose. We have opted for the second choice in order to generate a basic jointree which is as similar as possible to the original one. This is crucial for computation reuse as we shall discuss later.

[4]An instantiation of $\mathbf{E}$ contains one pair $(E, e)$ for each node $E$ in $\mathbf{E}$, where $e$ is a value of node $E$.

**Definition 7** *Let $\mathcal{G}$ be a dag and $\mathcal{T} = (\mathbf{V}, \mathbf{A}, \mathcal{H})$ be a basic jointree for $\mathcal{G}$ (induced by a spanning tree of its family graph). Let $q$ be a query for $\mathcal{G}$. The pruning of $\mathcal{T}$ given $q$, denoted $\mathcal{T}^q$, is a labeled tree $(\mathbf{V}, \mathbf{A}, \mathcal{H}^q)$ where*

$$\mathcal{H}_i^q \stackrel{def}{=} \begin{cases} \mathcal{H}_i, & \text{if } i \text{ is a node in } \mathcal{G}^q; \\ \emptyset, & \text{otherwise.} \end{cases}$$

That is, pruning a basic jointree is simply a matter of emptying the hypernode corresponding to each pruned node.

**Theorem 3** *In Definition 7, the pruned tree $\mathcal{T}^q$ is guaranteed to be a basic jointree for the pruned dag $\mathcal{G}^q$.*

Therefore, Definition 7 can be viewed as a proposal for reconfiguring a basic jointree in response to a query change. The reconfiguration suggested by this definition leaves the structure of the basic jointree intact, but it may change the contents of its hypernodes, therefore, leading to a possible change in separators and cliques.

We will close this section by discussing how inference is performed in this framework of dynamic jointrees. The first thing to observe is that in the jointrees we have been generating, each clique $\mathcal{C}_i$ has associated with it a single family; the family of node $i$. We will then associate the probability matrix of node $i$ with clique $\mathcal{C}_i$. Hence, initially, the local information associated with clique $\mathcal{C}_i$ is the matrix of node $i$ multiplied by any likelihood vector representing evidence available about node $i$. If node $i$ is pruned later, there will be no local information associated with its corresponding clique. For uniformity though, we assume that the local information is the unit potential $\mathbf{1}$ in such a case. From here on, we will use $\phi_i^q$ to represent the local information associated with clique $\mathcal{C}_i$ under query $q$. Moreover, for each edge $(i,j)$, we will use $\phi_{ij}^q$ to denote $\prod_k \phi_k^q$ where $k$ is a node on the $i$-side of edge $(i,j)$.

We can now define the message from node $i$ to node $j$ under query $q$ in the usual way:

$$\mathcal{M}_{ij}^q \stackrel{def}{=} \sum_{\mathcal{C}_i^q \setminus \mathcal{S}_{ij}^q} \phi_i^q \prod_{k \neq j} \mathcal{M}_{ki}^q. \qquad (1)$$

Note, here, that messages, separators and cliques have superscripts $q$ because they are now defined with respect to a particular query $q$ (and its pruned basic jointree $\mathcal{T}^q$).

To compute the probability distribution of a particular node $i$ under query $q$, we can then use the standard formula:

$$\mathbf{Pr}_i^q = \sum_{\mathcal{C}_i^q \setminus \{i\}} \phi_i^q \prod_k \mathcal{M}_{ki}^q. \qquad (2)$$

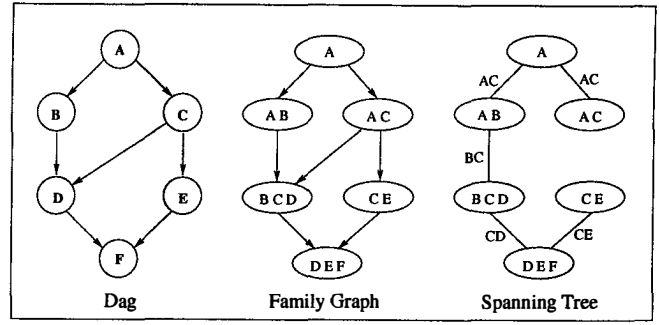

Figure 5: Computing a basic jointree as a spanning tree of the family graph.

Please note that Equations 1 and 2 are nothing but the standard algorithm for jointrees as given in [4, 5]. The only difference, however, is that everything now is indexed by a query $q$.

## 4 Reconfiguring a Jointree Efficiently

Reconfiguring a jointree in response to a query change involves two steps: (1) reconfiguring the underlying basic jointree; and (2) recomputing separators and cliques. To realize Step 1, all we have to do is decide which nodes have been pruned from the underlying dag given the current query and then update the hypernodes accordingly. Step 1 can be realized quite efficiently since pruning a belief network (according to Definition 6) can be done in time linear in the size of the network. Realizing Step 2, however, can be more involved since we must recompute $\mathcal{H}_{ij} \cap \mathcal{H}_{ji}$ for each edge $(i,j)$ in the basic jointree $(\mathbf{V}, \mathbf{A}, \mathcal{H})$. In the worst case, computing this intersection is quadratic in the number of nodes in the underlying dag.

One can do better than this, however, if one utilizes the following theorem.

**Theorem 4** *Let $(\mathbf{V}, \mathbf{A}, \mathcal{F})$ be a family graph and let $(\mathbf{V}, \mathbf{A}', \mathcal{F})$ be one of its spanning trees. Let*

$$\mathbf{S} \stackrel{def}{=} \{i : (i,j) \in \mathbf{A} \text{ and } (i,j) \notin \mathbf{A}' \text{ for some } j\}.$$

*For every edge $(i,j)$ in $\mathbf{A}$ and $\mathbf{A}'$, $\mathcal{F}_{ij} \cap \mathcal{F}_{ji} \setminus \{i\} \subseteq \mathbf{S}$.*

First, note that $\mathbf{S}$ is the set of all nodes that have lost outgoing edges in the process of rendering the family graph singly connected. In Figure 5 for example, only one node, $C$, has lost an outgoing edge in this process. Therefore, $\mathbf{S} = \{C\}$ in this case.

Theorem 4 is then saying that if a node $k$ is shared by families on opposite sides of edge $(i,j)$ (that is, $k \in \mathcal{F}_{ij} \cap \mathcal{F}_{ji}$), and if $k \neq i$, then it must be a node that has lost an outgoing edge in the process of rendering the family graph singly connected. The major implication of this theorem is:

**Corollary 1** *Regarding the edge $(i,j)$ in Theorem 4:*

$$\mathcal{S}_{ij} = \{i\} \cup (\mathbf{S} \cap \mathcal{F}_{ij} \cap \mathcal{F}_{ji}).$$

Consider Figure 5 again where $\mathbf{S} = \{C\}$. According to this corollary, the separator of any edge which extends from node $i$ to node $j$ can only contain node $i$ and, possibly, Node $C$.

Therefore, given Corollary 1, computing a separator is no longer quadratic in the number of nodes in the dag. It is only quadratic in the size of set $\mathbf{S}$; that is, it is only quadratic in the number of nodes that have lost outgoing edges in the process of rendering the family graph singly connected.

In fact, Corollary 1 suggests a heuristic for constructing jointrees from family graphs: minimize the number of nodes that will lose outgoing edges when rendering the family graph singly connected. Note that the size of any separator cannot exceed the size of set $\mathbf{S}$ plus 1; therefore, the previous heuristic will attempt to minimize the size of separators.[5]

We have taken advantage of the above theorem in our implementation, which has led to an efficient computation of separators. This has made the time spent on reconfiguring jointrees insignificant compared to the achieved savings, as our experimental results will demonstrate later.

## 5 Reusing Computations in Dynamic Jointrees

We now turn to the important issue of computation reuse. A key objective attempted by algorithms for belief network inference is to reuse computations across different queries. That is, having computed the posteriors of nodes $\mathbf{Q}$ given evidence $\mathbf{e}$, algorithms save the results of their intermediate computations for the possibility of reusing them when trying to compute the posteriors of nodes $\mathbf{Q}'$ given evidence $\mathbf{e}'$. In jointree algorithms, for example, if a message comes from a part of the jointree which did not involve a change in evidence, that message can be reused without having to recompute it again.

---

[5]An interesting implication of Corollary 4 is the following result. Compute a loop-cutset of the family graph and generate a spanning tree by eliminating (from the graph) only edges that are outgoing from nodes in the loop-cutset. This is always possible by definition of a loop-cutset. Under these conditions, we are guaranteed that no separator will have more than $c+1$ nodes where $c$ is the size of the loop-cutset.

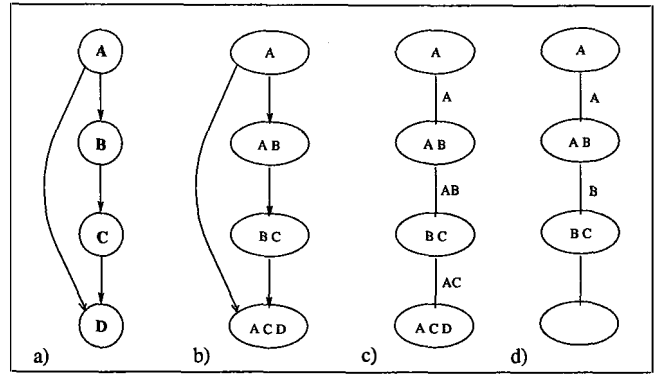

Figure 6: Figures (c) and (d) depict basic jointrees for the queries $(\{\}, \{D\})$ and $(\{\}, \{C\})$, respectively.

In this section, we start by a theorem which states conditions under which a message computed with respect to a basic jointree $\mathcal{T}^{q_1}$ will remain valid (can be reused) in the reconfigured basic jointree $\mathcal{T}^{q_2}$. Our results are with respect to a dag $\mathcal{G}$, its basic jointree $\mathcal{T}$, and two corresponding queries $q_1$ and $q_2$.

**Theorem 5** *For every edge $(i,j)$, if $\phi_{ij}^{q_1} = \phi_{ij}^{q_2}$ and $\mathcal{S}_{ij}^{q_2} = \mathcal{S}_{ij}^{q_1}$, then $\mathcal{M}_{ij}^{q_2} = \mathcal{M}_{ij}^{q_1}$.*

That is, if the local information associated with nodes on the $i$-side of edge $(i,j)$ did not change when switching from query $q_1$ to query $q_2$, then the message from node $i$ to node $j$ can be reused. This is similar to message reuse in standard jointree algorithms except that standard algorithms do not check for the condition $\mathcal{S}_{ij}^{q_2} = \mathcal{S}_{ij}^{q_1}$; the condition always hold given that the jointree does not change.

We do have, however, an even stronger theorem which implies the one given above.

**Theorem 6** *For every edge $(i,j)$, if $\phi_{ij}^{q_1} = \phi_{ij}^{q_2}$ and $\mathcal{S}_{ij}^{q_2} \subseteq \mathcal{S}_{ij}^{q_1}$, then*

$$\mathcal{M}_{ij}^{q_2} = \sum_{\mathcal{S}_{ij}^{q_1} \setminus \mathcal{S}_{ij}^{q_2}} \mathcal{M}_{ij}^{q_1}.$$

That is, even if the separator corresponding to a particular message does change as a result of changing the query, it may still be possible to reuse that message but only after marginalizing it.

Let us consider an example illustrating the use of Theorem 6. Figure 6(a) depicts a belief network and Figure 6(b) depicts its family graph. Suppose now that we have the query $q_1 = (\{\}, \{D\})$: we are interested in the prior distribution of Node $D$. The basic jointree in Figure 6(c) can be used to answer this query. Using Equations 1 and 2 to compute the distribution of Node $D$, we end up computing three messages: $\mathcal{M}_{AB}$, $\mathcal{M}_{BC}$ and $\mathcal{M}_{CD}$. Suppose now that the query changes to $q_2 = (\{\}, \{C\})$: we are now interested in the prior distribution of Node $C$. Node $D$ becomes irrelevant and is pruned given this new query. Reconfiguring the basic jointree in Figure 6(c), we obtain the one in Figure 6(d). Note here the change in separators. In particular, the separator between Nodes $B$ and $C$ has changed from $\{A, B\}$ to $\{B\}$. Since the local information associated with Nodes $A$ and $B$ remain the same, Theorem 6 tells us that

$$\mathcal{M}^{q_2}_{BC} = \sum_{\{A\}} \mathcal{M}^{q_1}_{BC}.$$

That is, there is no need to recompute the message from Node $B$ to Node $C$, we can simply reuse its previous value after we have marginalized it.

## 6 Preliminary Experimental Results

If there are no query changes, our method will degenerate into the standard method for computing with jointrees, except possibly for the fact that we are constructing jointrees differently. However, if the query changes, then our method will reconfigure the jointree in response, reusing some — but not necessarily all — computations that have been performed with respect to the previous jointree.

Our goal in this section is twofold. First, to substantiate the claim that reconfiguring a jointree (as accomplished by our method) consumes insignificant time compared to the achieved savings. Second, to give an indication of the amount of saving possible as a result of reconfiguring a jointree.

We start by describing our experiments, that is, how we generated our belief networks and the corresponding queries. Each belief network was generated given three parameters: the number of nodes, a probability distribution over sizes of families, and a graph-width parameter which controls the connectivity of generated networks. Each of the generated networks had, on average, 20% root nodes, 10% single-parent nodes, 25% two-parent nodes, 35% three-parent nodes and 10% four-parent nodes. The graph-width parameter was varied to generate networks with separators containing up to 20 nodes.

We have conducted two sets of experiments. In the first, we attempted a standard computation: Compute the prior distribution of each leaf node in a belief network (no evidence). However, we did this as follows. Given leaf nodes $1, 2, \ldots, i$, we have generated $i$ queries $(\{\}, \{1\}), (\{\}, \{2\}), \ldots, (\{\}, \{i\})$. This has forced our algorithm to reconfigure the jointree before each query is attempted. We have counted the number of additions and multiplications performed by our algorithm to compute these priors. We have also counted the same number of operations by running the algorithm *but without reconfiguring the jointree.*

Table 1 depicts the results of this experiment. It shows ten sets of networks, each set containing 50 networks generated randomly using the same parameters. For each of the sets, the table shows the average and maximum saving factor: number of operations for the static jointree algorithm divided by the number of operations for the dynamic jointree algorithm. Note that for certain networks, the saving factor is as large as 282.54! In fact, it appears that the more connected the network is, the higher the saving factor is. Table 1 supports this observation by showing the average size of the maximal separator for both dynamic and static jointrees. On average, it is clear that reconfiguring jointrees leads to reducing the size of the maximal separator.

Table 1 also lists (in the last column) the ratio between the time spent reconfiguring jointrees and the time spent on standard inference — this is shown as a percentage. It should be clear from this table that this percentage is insignificant when viewed in light of the saving entailed by jointree reconfiguration. Again, notice how this percentage gets smaller as the networks get more connected.

The second experiment involved changing evidence only. For this experiment, we picked up 10% of a network's non-root nodes, instantiated them randomly, and then computed the posterior distribution for each root node in the network. We then considered each of the instantiated nodes in turn, changing the evidence on it randomly and recomputing the posterior distribution of each root node. We repeated this process five times for each network.

Our goal here was to generate a mixture of evidence changes, some leading to significant jointree reconfiguration and others leading to no reconfiguration. Each of the five repetitions per network leads to a significant jointree reconfiguration since evidence nodes are changing significantly. Within each repetition, however, no reconfiguration takes place since evidence nodes are the same — only their observed values change.

Table 2 shows ten sets of networks, each set containing 50 networks generated randomly using the same parameters. The table shows the same indicators shown in Table 1. Note that Table 2 supports the same hypotheses supported by Table 1: The saving factor increases as the networks become more connected; The time spent on reconfiguring jointrees is insignificant relative to the achieved savings; Dynamic reconfigura-tion of jointrees does reduce the size of maximal separator on average; The savings due to reconfiguring jointrees can be quite significant, getting close to a factor of 300 in certain situations.

Table 1: Evaluating the savings entailed by dynamically reconfiguring jointrees.

| Network | Saving Factor | | Average Size of Maximal Separator | | Reconfiguration |
| Size | Average | Maximum | Dynamic JT | Static JT | Time % |
| --- | --- | --- | --- | --- | --- |
| 50 | 1.44 | 2.20 | 5.48 | 5.74 | 43.10 |
| 50 | 1.85 | 4.14 | 6.58 | 7.24 | 21.06 |
| 50 | 2.98 | 9.81 | 7.88 | 9.24 | 18.00 |
| 50 | 4.17 | 19.35 | 8.26 | 9.76 | 10.30 |
| 50 | 7.76 | 36.90 | 8.90 | 11.04 | 13.04 |
| 75 | 1.69 | 4.24 | 6.90 | 7.56 | 26.90 |
| 75 | 3.05 | 10.91 | 8.68 | 9.98 | 13.68 |
| 75 | 5.21 | 20.74 | 9.70 | 11.68 | 7.78 |
| 75 | 11.95 | 65.80 | 9.96 | 13.12 | 6.96 |
| 75 | 27.64 | 282.54 | 11.06 | 14.54 | 5.54 |

Table 2: Evaluating the savings entailed by dynamically reconfiguring jointrees.

| Network | Saving Factor | | Average Size of Maximal Separator | | Reconfiguration |
| Size | Average | Maximum | Dynamic JT | Static JT | Time % |
| --- | --- | --- | --- | --- | --- |
| 50 | 2.99 | 13.8 | 5.70 | 6.20 | 5.88 |
| 50 | 5.00 | 14.18 | 6.30 | 7.36 | 4.94 |
| 50 | 11.51 | 74.63 | 7.56 | 8.92 | 4.74 |
| 50 | 16.16 | 165.20 | 8.08 | 9.82 | 4.66 |
| 50 | 22.75 | 250.58 | 8.64 | 10.46 | 3.24 |
| 75 | 4.41 | 18.16 | 6.88 | 7.48 | 5.68 |
| 75 | 13.01 | 81.75 | 8.12 | 9.96 | 3.00 |
| 75 | 26.43 | 181.95 | 9.88 | 12.04 | 2.40 |
| 75 | 42.11 | 289.88 | 10.76 | 13.68 | 1.76 |
| 75 | 56.58 | 294.02 | 11.16 | 14.24 | 1.64 |

The reported experiments are by no means comprehensive. However, they involve a total of 1000 networks with many queries attempted per network. Their indications, therefore, are not to be underestimated.

## 7 Conclusion

This paper is based on two contributions. First, a non-classical definition of jointrees which stresses properties of jointrees that have not been given enough attention in the literature. Second, an application of this non-classical definition to inference situations involving considerable query changes.

For these kind of situations, we have proposed to reconfigure the jointree as the query changes. We have also proposed a specific method for this reconfiguration and shown that it satisfies two important properties. First, it can be done efficiently. Second, it allows the reuse of some results computed before the reconfiguration takes place.

Finally, we provided a preliminary experimental analysis indicating that significant savings can be expected from reconfiguring jointrees in situations where query changes are considerable.